# A Novel Retinal Vessel Segmentation Based On Histogram Transformation Using 2-D Morlet Wavelet and Supervised Classification

Saeid Fazli
Research Institute of Modern Biological Techniques
University of zanjan
Zanjan, Iran
fazli@znu.ac.ir

Sevin Samadi
Electrical Engineering Department
University of zanjan
Zanjan, Iran
Samadi_sevin@znu.ac.ir

*Abstract* —The appearance and structure of blood vessels in retinal images have an important role in diagnosis of diseases. This paper proposes a method for automatic retinal vessel segmentation. In this work, a novel preprocessing based on local histogram equalization is used to enhance the original image then pixels are classified as vessel and non-vessel using a classifier. For this classification, special feature vectors are organized based on responses to Morlet wavelet. Morlet wavelet is a continues transform which has the ability to filter existing noises after preprocessing. Bayesian classifier is used and Gaussian mixture model (GMM) is its likelihood function. The probability distributions are approximated according to training set of manual that has been segmented by a specialist. After this, morphological transforms are used in different directions to make the existing discontinuities uniform on the DRIVE database, it achieves the accuracy about 0.9571 which shows that it is an accurate method among the available ones for retinal vessel segmentation.

## I. INTRODUCTION

Retinal angiography images are extensively used in the diagnosis of important diseases such as hypertension, arteriosclerosis and diabetes. Furthermore, there has been an increasing inclination for personal authentication techniques by using human biometric features Retinal is one of these features.

One of the most important diseases that causes blood vessels structure to change is diabetic retinopathy that leads to adults blindness. To overcome this problem specialist analysis is required [1].valuable applications have been described in [2, 3] for identification of a variety of systematic diseases such as diabetes and hypertension. Several supervised [2-4] and unsupervised [5-7] vessel segmentation methods have already been proposed and implemented. In the literature, several techniques have been reported for blood vessel segmentation and diagnosis of such diseases [8–13]. These methods generally can be classified into three categories: (1) kernel-based, (2) tracking based and (3) classifier-based methods. Kernel-based methods convolve the image with a kernel based on a predefined model [14, 15]. Methods based on classifier include two steps: first, a segmentation of the image and then a classification of regions. In tracking based methods vessels edge are followed using local information.

There are also other methods in which two of above methods are combined [16, 18]. For image analysis, detecting the vessels means generating a binary mask that helps us to label pixels as vessel or background. The goal is to find and detect more details, at the same time to avoid false positives and, ideally, to keep vessel connectivity. It should be noticed that many clinical studies do not use fine vessels, just taking measurements on main ones in area around optic disc [19, 20, 21].

Traditional approaches to retinal vessel segmentation mostly use line detection and tracking based methods [32].Since line detection methods are reliant and do not have acceptable result in all cases [11]. A side from these traditional methods retinal vessel segmentation using classifiers has become popular recently [32].

In previous works, continuous wavelet transform (CWT) [22,23] is used and for the next step, integration of multi-scale information is used for supervised classification [24].Here a Bayesian classifier is used with Gaussian mixture models as class likelihoods and evaluate performances by accuracy analysis.

In pixel classification approach [33], a feature vector is constructed for each pixel of the image and two-dimensional Gabor wavelet transform responses taken at multiple scales. Then, a classifier is trained using these feature vectors to segment the image. Each pixel is represented by a feature vector composed of the pixel's intensity and two-dimensional Gabor wavelet transform responses taken at multiple scales.

In our approach, each pixel is represented by a feature vector which uses measurements of different scales taken from Morlet wavelet transform. A Bayesian classifier with class conditional probability density functions is used and described as Gaussian mixtures that can model complex decision surface.

The CWT is a robust transform and has been applied to many different image processing problems, from image coding [25] to shape analysis [26]. Wavelets are especially suitable for detecting singularities in images [27], extracting immediate frequencies [28], and performing fractal and multi-fractal analysis. The Morlet wavelet can adjust to different frequencies and it can reduce noise in single step. These characteristics make it suitable for our work.

This paper proposes a novel method to segment blood vessels. The proposed algorithm contains four steps as follows:

- A novel vessel enhancement technique based on the histogram of gray scaled image and dividing original image in to blocks is introduced. Local adaptive histogram equalization is used for each block.
- Feature generation process is done using Morlet wavelet and comparing it with 2-D CWT.
- A Bayesian classifier with Gaussian mixture models as class likelihoods is used for classification of pixels.
- In the last step, a post processing is done which includes two steps. At first a median filter is utilized for enhancing the hidden pixels which belong to vessel and simultaneously removes noisy pixels. Although this step improves the output, still some error is remaining in classification. After median filter, morphological transforms are applied to image in five different directions. The output image of this process is derived using the logical OR of the five responses.

## II. MATERIALS AND METHODS

### A. Materials

Our methods are tested and evaluated on publicly available database of non-mydriatic images and corresponding manual segmentations: the DRIVE [29].

The DRIVE database consists of 40 images including their manual segmentation by experts. These images are captured in digital form from a canon CR5 non-mydriatic 3CCD camera in a field of view. The images are of size 768*584 pixels, 8 bit per color. All images are in JPEG format.

The 40 images have been divided into training and test sets, each containing 20 images. They have been manually segmented by three observers trained by an ophthalmologist. The images in the training set were segmented once, while images in the test set were segmented twice, resulting in sets A and B. The observers of sets A and B produced similar segmentations. In set A, 12.7% of pixels where marked as vessel, against 12.3% vessel for set B. Performance was measured on the test set using the segmentations of set A as ground truth. The segmentations of set B were tested against those of A, serving as a human observer reference for performance comparison [5].

### B. Pre processing

In retinal images, there is an insignificant difference between intensity level of vessels and non-vessels regions, consequently intensity transform is used for improving algorithms result. There are several transforms for this usage such as Histogram Equalization (HE). HE uses histogram of original image and convert it to image with uniform histogram. However, this transform is not so successful in our experiments. It can improve the visual appearance of images.

Adaptive histogram equalization is a method that used the HE in multiple local window size area emphasizes local contrast, rather than overall contrast. AHE algorithms find local mappings using local histograms. Recently many researches in the medical field have begun using AHE as an enhancement method for various diagnosis using radiographs images [34, 35].

In this paper, at first the image is changed to grayscale. Secondly, local transformation with different algorithms is used on our gray scaled image. In the second part we came to the conclusion that using local adaptive histogram is the best intensity transform for this purpose. After this pre processing and making better the images intensity, these developed images are exposed to the algorithm and a satisfactory change in accuracy rate is seen.

AHE is the histogram of gray levels (GL's) in which a window around each pixel is generated. Then the distribution of GL's is increased and cumulative sum over the histogram is calculated. In all the processes the input image is compared with the output image. If a pixel has GL lower than others in the surrounding window, the output is black; if it has median value in its window, the output is 50% gray.

The procedure of preprocessing is show in Fig1.

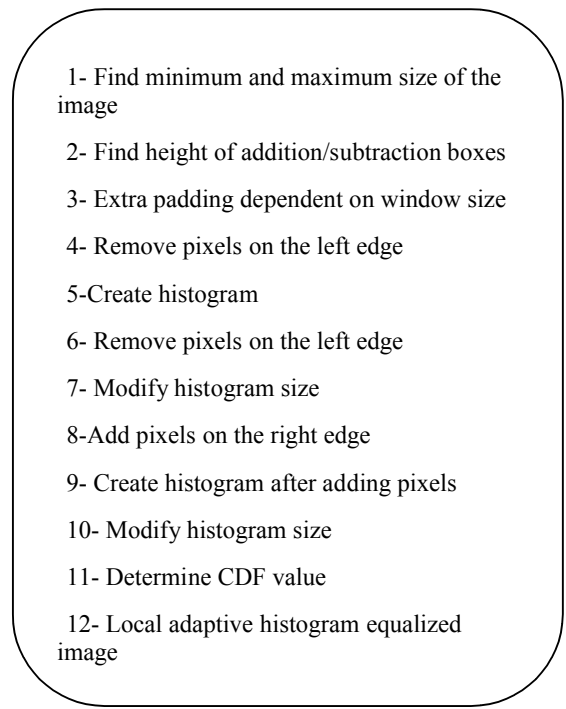

1- Find minimum and maximum size of the image

2- Find height of addition/subtraction boxes

3- Extra padding dependent on window size

4- Remove pixels on the left edge

5- Create histogram

6- Remove pixels on the left edge

7- Modify histogram size

8- Add pixels on the right edge

9- Create histogram after adding pixels

10- Modify histogram size

11- Determine CDF value

12- Local adaptive histogram equalized image

Figure 1: the procedure of method

The result is an image in which the mapping applied to each pixel is different and is adaptive to the local distribution of pixel intensities rather than the global information content of the image. In practice, this produces an image in which different objects whose intensity values lie in different sub-ranges of the intensity values are simultaneously visible. It is clear that AHE allows the simultaneous visualization of the major vessels and thin vessels.

After all procedures done on the image, the local adaptive histogram image is available and the algorithms can be tested on that.

Differences between original image and the preprocessed one is shown in fig2. As it can be seen vessels are more vivid.

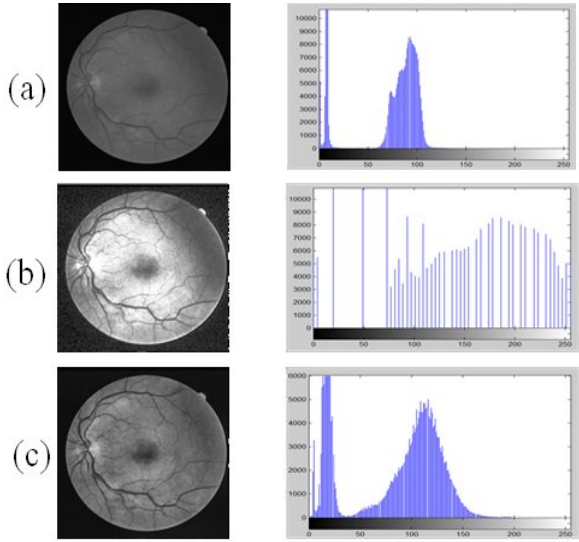

Figure 2. (a) gray scaled original image and histogram (b) histogram equalized image (c) local adaptive histogram equalization

### C. Features Selection

Wavelet and curvelet are multiscale transforms. They are recognized as useful feature extraction methods to represent image features. They have a great performance in detection point and line features.

The continuous wavelet transform is defined as:

$$T\psi(b,\theta,a) = C_\psi^{-1/2} 1/a \int \psi^*(a^{1-}r_{-\theta}(x-b))f(x)d^2x \quad (1)$$

Where $C\psi$, $\psi$, b, θ and a, denote the normalizing constant, analyzing wavelet, the displacement vector, the rotation angle and the dilation parameter (also known as scale), respectively $\psi$ denotes the complex conjugate of $\psi$ [29].

The Morlet wavelet and Gabor transform have lots of similarities but still some differences remain. Because of these diversities Morlet transform is selected for this framework. One of the most important differences that encourage us using Morlet is the flexible window size with scaling parameter while the size of window in Gabor transform is fixed.

There are several types of wavelets, such as 2-D Mexican hat and the optical wavelet but 2-D Morlet wavelet is chosen. This type of wavelet is suitable for our purpose since it has the capability of detecting oriented features and tuning to specific frequencies. Since it can adjust to the frequency, background noise can be removed. The 2-D Morlet wavelet is defined as:

$$\psi_M(x) = \exp(jk_0 x)\exp(-\frac{1}{2}|Ax^2|) \quad (2)$$

Where $j = \sqrt{-1}$ and $A = diag[\varepsilon^{-1/2}, 1], \varepsilon \geq 1$ is a 2×2 diagonal matrix that defines the anisotropy of the filter. The Morlet wavelet is a complex exponential modulated Gaussian function.

The aim is to find maximum modulus in all scales and over all orientations. To achieve this goal, wavelet is computed for degrees between 0 up to 170 at step of 10 degrees and the maximum is chosen.

Features are defined dimensional, and so due to this description Error rate might be increased. To solve this problem features are normalized and the normalization transform is defined as:

$$\hat{v}_i = \frac{v_i - \mu_i}{\sigma_i} \quad (3)$$

Where $v_i$ is the $i_{th}$ feature assumed by each pixel, $\mu_i$ is the average value of the $i_{th}$ feature and $\sigma_i$ is the associated standard deviation. After the normalization process, they are applied to all features; each feature space is normalized by its own mean value.

### D. Classification for Segmentation

Different scales of Morlet transform allow us to detect vessels with various thicknesses. The pixels of image are viewed as objects that are represented by feature vectors so statistical classifiers can be applied for segmentation. This approach provides requirements to combine information from wavelet responses at multiple scales to distinguish pixels from each class [5].

Bayesian classification is based on probability theory and the principle of choosing the most probable or the lowest risk option [31]. Since the blood vessel has line structure and it can be modeled as Gaussian function wavelet analysis with Gaussian kernel is used.

Supervised classification has been applied for the final segmentation. It is assigning any pixel to vessel and non-vessel. To obtain the final segmented image supervised classification is used, two different classes are defined, $C_1$ for vessel pixels and $C_2$ for non-vessels. Bayesian classifier is used for this purpose and GMM (A Bayesian classifier in which each class-conditional probability density function is described as a linear combination of Gaussian functions) is defined as model. For all classifications a training set is needed that can be a part of segmented image or a set of pixels which is classified by an expert. In this case, a random subset of segmented images by an expert is used [5].

The Gaussian probability density is a bell shaped curve that described by two parameters, mean and variance. Fig 3 shows an example of 2d-Gaussian PDF.

The Gaussian distribution is usually quite good approximation for a class model shape in a suitably selected feature space. It is mathematically used for one dimensions but it can easily extended to multiple dimension.

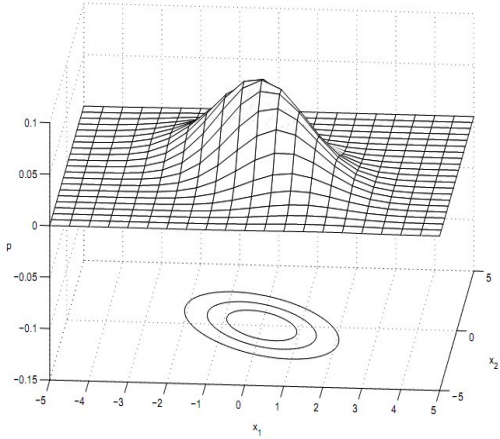

Figure 3 Surface of 2d- Gaussian PDF

Gaussian Mixture Model is a method between non-parametric and parametric models, providing a relatively fast classification process at the cost of a more expensive training algorithm.

The Gaussian Mixture Model Classifier (GMM) is a useful supervised learning classification algorithm that can be used to classify a wide variety of N-dimensional signals. Since a Bayesian classifier is used, linear combination of Gaussian functions are defined as probability function. A feature vector is said to belong to a class if it maximizes:

$$p(C_i|x_t) = p(x_t|C_i)p(C_i) \qquad (4)$$

In the case where all the classes are assumed and can occur with the same probability, we are actually concerned by maximizing for every possible class.

Probabilities are defined as follow in which vector D is the feature vector:

$$\begin{array}{l} C_1 \text{ if } p(C_1|D) > p(C_2|D) \\ \text{Otherwise } C_2 \end{array} \qquad (5)$$

Bayes rule is defined as:

$$p(C_i|D) = \frac{p(D|C_i)p(C_i)}{p(D)} \qquad (6)$$

Where $p(C_i)$ is the prior probability of class $C_i$ and $p(D)$ is the probability density function of $D$.

Using Bayes rule and combining it with probability conditions help us to obtain the best result. New condition which is combined with this rule is defined as:

$$\begin{array}{l} C_1 \text{ if } p(C_1|D)p(C_1) > p(C_2|D)p(C_2) \\ \text{Otherwise } C_2 \end{array} \qquad (7)$$

The class likelihoods are described as linear combinations of Gaussian functions:

$$p(D|C_i) = \sum_{j=1}^{k_i} p(D|j,C_i)P_i \qquad (8)$$

Where $k_i$ is the number of Gaussians modeling likelihood $p_j$ is the weight of Gaussian, j in $p(D|j,C_i)$ is a d-dimensional Gaussian distribution. GMM provides a fast classification process in the complex and expensive training algorithm.

After this classification, still some parts are not connected for connecting these parts and removing noise median filter is used.

After applying the median filter still some mis-classifications can be detected. To eliminate this error, Five different morphological openings in five directions 0°, 30°, 60°, 120° and 150° are used [30]. After implementation of this process length of vessels must be considered to retain only vessel like structures. In this part objects with length equal or larger than desired one is preserved. Output of this process is gained after a logical OR between different outputs.

For the DRIVE database, 20 labeled training images are used as training set. These images are segmented by an expert and used for evaluation of retinal vessel segmentation algorithms.

For making this more clear in the following part one of the DRIVE database images is selected to make the differences more evident. Fig.4 (a) shows an original image from DRIVE database. Fig.4 (b) shows the image after local histogram analysis. Fig.4(c) shows the probabilities of the image. Fig.4 (d) shows the image after classification using GMM. Fig.4 (e) shows the image after median filtering. Fig.4 (f) shows the image after post processing. Fig.4 (g) is the ground truth of this image.

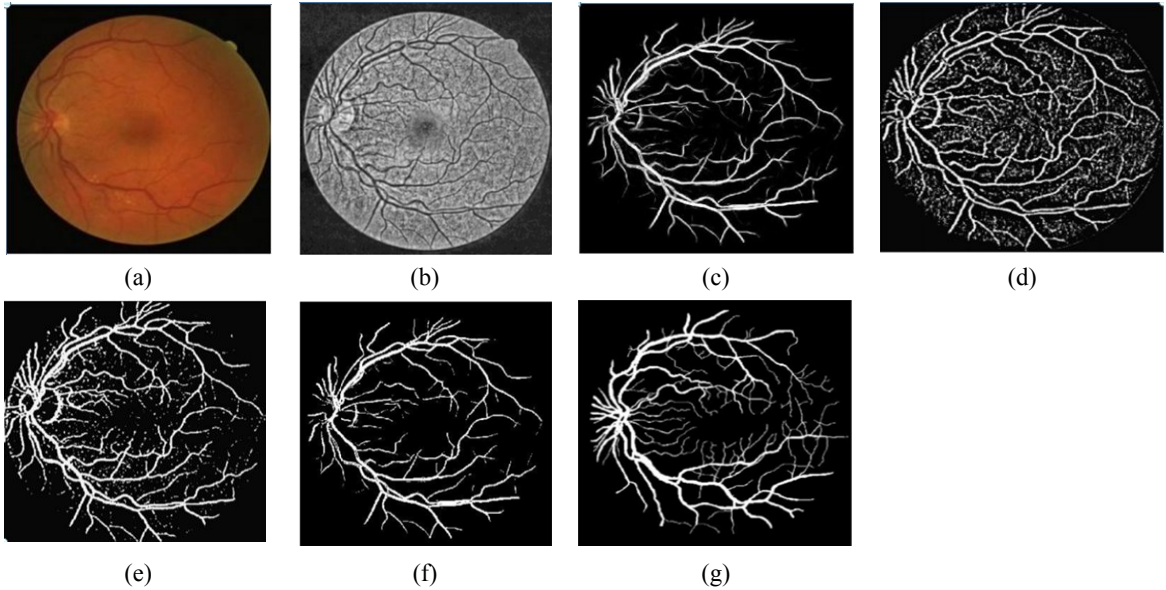

(a) (b) (c) (d)

(e) (f) (g)

Figure 4: (a)original image (b)histogram equalized image (c) image's probabilities (d) after classification using the GMM (e) median filtering (f)shows the image after post processing(g) is the ground truth

## III. EXPERIMENTAL RESULTS

The proposed algorithm is evaluated using the retinal images of DRIVE database, containing both healthy people and those suffering from diabetic retinopathy. In this database, 40 images are randomly selected between 400 images. 33 of them belong to healthy retinal and the rest have signs of diabetic retinopathy.

The Morlet transform enhance the vessel contrast and filter out the noise. It is used in different scales and makes it possible to segment vessels of different orientations. Using manual for learning allows the approach to be trained for different type of images.

In retinal vessel segmentation, at last we reach a result that is a classification based on pixels. Each pixel is categorized as vessel or non-vessel. To evaluate the outcome of the process, four classes of pixels must be investigated, true positive (TP) and true negative (TN) when a pixel is correctly segmented as a vessel or non-vessel, and two misclassifications, a false negative (FN) appears when a pixel in a vessel is segmented in the non-vessel area, and a false positive (FP) when a non-vessel pixel is segmented as a vessel-pixel.

Two widely known measurements are used for evaluation of this method: sensitivity and specificity. These performance measures were defined and widely used in literature [5, 30, 37].Sensitivity is a normalized measure of true positives, while specificity measures the proportion of true negatives:

$$\text{Sensitivity} = \frac{TP}{TP+FN}$$

$$\text{Specificity} = \frac{TN}{TN+FP}$$

The accuracy of the binary classification is defined by

$$\text{Accuracy} = \frac{TP+TN}{P+N}$$

Where P and N represent the total number of vessel and non-vessel pixels in the segmentation process. The accuracy shows the degree of conformity between the output and the manual of original image. Thus, the accuracy is strongly related to the segmentation property and shows how proper are the segmentation method. For this reason it is used to evaluate and compare different methods.

Table.1 shows the algorithms result and compare it with others.

TABLE 1 Accuracy of different algorithms

| Vessel detection methods | Average accuracy |
|---|---|
| Gaussian match filter [31] | 0.8850 |
| Niemeijer[32] | 0.9416 |
| Jiang[33] | 0.9327 |
| Chuadhari[34] | 0.9103 |
| Mf/ant[35] | 0.9293 |
| This work | 0.9571 |

The tradeoff between the parameters measurement is represented with the receiver operating characteristic curve (ROC), which is a plot of the sensitivity versus 1−specificity. Equivalently, ROC curve can be represented by plotting the true positive rate (TPR) versus the false positive rate (FPR) [36].

$$TPR = \frac{TP}{TP+FN}$$

$$FPR = \frac{FP}{FP+TN}$$

In fig 5, the ROC curve is shown. After comparing the results with manual, true and false positive fractions are provided. Using these rates, ROC curve can be plotted. The closer ROC curve to top left corner, the better the performance of the approach. The areas under the ROC curves are used to measure the performance of each approach.

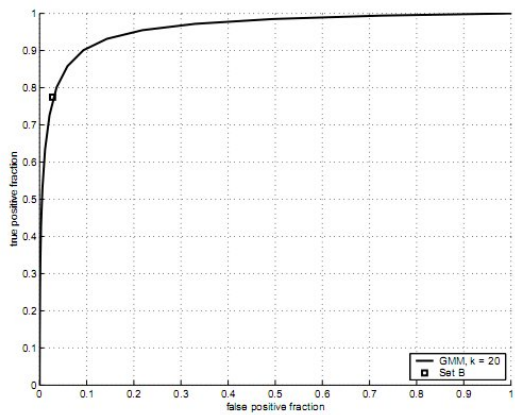

Figure 5 Roc of the proposed method

## IV. CONCLUSION

In this paper, novel retinal vessel segmentation algorithm is proposed. A new pre processing step based on local histogram equalization is introduced and GMM classifier is used for classification. After the post processing and making some changes in classifiers parameters, the output is obtained. Comparing the results the proposed algorithm to some other existing ones indicates the effectiveness of the proposed method as shown in Table 1. Also supervised classification method minimizes the users interaction. The experimental results reveal that it considerably reduces the false detection. The proposed technique can be used in both healthy and unhealthy retinal vessel segmentation. The performance of this method is shown by accuracy measurements on DRIVE database. This approach increases the accuracy rate about 2 percent compared to the existing methods.

REFERENCES

[1] J. Sussman, W. G. Tsiaras, and K. A. Soper, "Diagnosis of diabetic eye disease," Journal of the American Medical Association, vol. 247,pp. 3231–3234, 1982

[2] Li H, Chutatape O. Automated feature extraction in color retinal images by a model based approach. IEEE Trans Biomed Eng 2004;51(2):246–54.

[3] Walter T, Kein JC, Massin P, Erginay A. A contribution of image process-ing to the diagnosis of diabetic retinopathy—detection of exudates in color fundus images of the human retina. IEEE Trans Med Imaging 2002;21(10):1236–43

[4] Leandro JJG, Cesar Jr RM, Jelinek H. Blood vessels segmentation in retina: preliminary assessment of the mathematical morphology and of the wavelet transform techniques. In: Proceedings of the 14th Brazilian symposium on computer graphics and image processing. 2001. p. 84–90.

[5] Soares JVB, Leandro JJG, Cesar Jr RM, JelinekHF, Cree MJ. Retinal vessel segmen-tation using the 2D Morlet wavelet and supervised classification. IEEE Trans Med Imaging 2006;25(9):1214–22

[6] Staal J, AbramoffMD, NiemeijerM, ViergeverMA, vanGinnekenB. Ridge-based vessel segmentation in color images for the retina. IEEE Trans Med Imaging 2004;23(4):501–9.

[7] Chaudhuri S, Chatterjee S, Katz N, Nelson M, Goldbaum M. Detection of blood vessels in retinal images using two dimensional matched filters. IEEE Trans Med Imaging 1989;8(3):263–9.

[8] ChutatapeO, Zheng L, Krishnan SM. Retinal blood vessel detection and tracking by matched Gaussian and Kalman filters. In: Proceedings of the 20th annual international conference of the IEEE on engineering in medicine and biology, vol. 6. 1998. p. 3144–9.

[9] Chanwimaluang T, Guoliang F. An efficient blood vessel detection algorithm for retinal images using local entropy thresholding. Proc Int Symp Circuits Syst (ISCAS) 2003;5:21–4

[10] S. Chaudhuri, S. Chatterjee, N. Katz, M. Nelson, M. Goldbaum, Detection of blood vessels in retinal images using two-dimensional matched filters, IEEE Transactions on Medical Imaging 8 (3) (1989) 263–269.

[11] A. Hoover, V. Kouznetsova, M. Goldbaum, Locating blood vessels in retinal images by piecewise threshold probing of a matched filter response, IEEE Trans-actions on Medical Imaging 19 (3) (2000) 203–210.

[12] M. Niemeijer, J.J. Staal, B. VanGinneken, M. Loog, M.D. Abramoff, Compara-tive study of retinal vessel segmentation methods on a new publicly available database, SPIE Medical Imaging 5370 (2004) 648–656.

[13] J. Ng, S.T. Clay, S.A. Barman, A.R. Fielder, M.J. Moseley, K.H. Parker, C. Paterson, Maximum likelihood estimation of vessel parameters from scale space analysis, Image and Vision Computing 28 (2010) 55–63.

[14] S.R. Nirmala, S. Dandapat, P.K. Bora, Wavelet weighted blood vessel distortion measure for retinal images, Biomedical Signal Processing and Control 5 (4) (2010) 282–291

[15] B.S.Y. Lam, Y. Gao, A.W.C. Liew, General retinal vessel segmentation using regularization-based multiconcavity modeling, IEEE Transactions on Medical Imaging 29 (7) (2010) 1369–1381.

[16] L. Zhou, M. Rzeszotarski, L.Singerman, and J. Chokreff, "The detection and quantification of retinopathy using digital angiograms,"IEEE Trans. Med. Imag., vol. 13, pp. 619–626, 1994

[17] X. Jiang and D. Mojon, "Adaptive local thresholding by verification based multithreshold probing with application to vessel detection in retinal images,"IEEE Trans. Pattern Anal. Mach. Intell., vol. 25, pp.

[18] A. A. Mendosnça and A. Campilho, "Segmentation of retinal blood vessels by combining the detection of centerlines and morphological reconstruction," IEEE Trans. Med. Imag., vol. 25, no. 9, pp.1200–1213, Sep. 2006

[19] M. E. Martinez-Perez, A. D. Hughes, S. Thom, A. A. Bharath, and K.H. Parker, "Segmentation of blood vessels from red-free and fluorescein retinal images," Med. Image Anal., vol. 11, pp. 47–61, 2007.

[20] N. Patton, T. Aslam, T. MacGillivray, A. Pattie, I. Deary, B. Dhillon, Retinal vascular image analysis as a potential screening tool for cerebrovascular disease: a rational based on homology between cerebral and retinal micro vasculatures, Journal of Anatomy 20 (2005) 319–348

[21] J. Ding, N. Patton, I. Deary, M. trachan, F. Fowkes, R. Mitchell, J. Price, Retinal microvascular abnormalities and cognitive dysfunction: a systematic review, British Journal of Ophthalmology 92 (2008) 1017–1025.

[22] T. Wong, M. Knudtson, R. Klein, S. Meuer, L. Hubbard, Computer-assisted measurement of retinal vessel diameters in the Beaver Dam

Eye Study: methodology, correlation between eyes, and effect of refractive errors, Journal of Ophthalmology 11 (2004) 1183–1190

[23] J. J. G. Leandro, R. M. Cesar-Jr., and H. Jelinek, "Blood vessels segmentation in retina: Preliminary assessment of the mathematical morphology and of the wavelet transform techniques," in Proc. of the 14th Brazilian Symposium on Computer Graphics and Image Processing. IEEEComputer Society, 2001, pp. 84–90.

[24] H. F. Jelinek and R. M. Cesar-Jr., "Segmentation of retinal fundus vasculature in non-mydriatic camera images using wavelets," in Angiography and Plaque Imaging: Advanced Segmentation Techniques, J. Suri and T. Laxminarayan, Eds. CRC Press, 2003, pp. 193–224

[25] J. J. G. Leandro, J. V. B. Soares, R. M. Cesar-Jr., and H. F . Jelinek,"Blood vessels segmentation in non-mydriatic images using waveletsand statistical classifiers," in Proc. of the 16th Brazilian Symposium on Computer Graphics and Image Processing . IEEE Computer Society Press, 2003, pp. 262–269

[26] O. Rioul and M. Vetterli, "Wavelets and signal processing," IEEE Signal Processing Magazine , pp. 14–38, Oct. 1991.

[27] L. da F. Costa and R. M. Cesar-Jr., Shape analysis and classification: theory and practice . CRC Press, 2001

[28] A. Grossmann, "Wavelet transforms and edge detection, " in Stochastic Processes in Physics and Engineering , S. A. et al., Ed. D. Reidel Publishing Company, 1988, pp. 149–157

[29] J.-P. Antoine, P. Carette, R. Murenzi, and B. Piette, "Image analysiswith two-dimensional continuous wavelet transform," Signal Processing , vol. 31, pp. 241–272, 1993

[30] J. J. Staal, M. D. Abr`amoff, M. Niemeijer, M. A. Viergever, and B. van Ginneken, "Ridge based vessel segmentation in color images of the retina," IEEE Transactions on Medical Imaging , vol. 23, no. 4, pp. 501–509, 2004

[31] S. Theodoridis and K. Koutroumbas. Pattern Recognition. Academic Press, 1999. ISBN 0-12-686140-4.

[32] R.Ghaderi H.Hassanpour M.Shahiri, "Retinal Vessel Segmentation Using the 2-D Morlet Wavelet and Neural Network". International Conference on Intelligent and Advanced Systems 2007.

[33] N. Armande,, P. Montesinos. "Thin nets and crest lines:Application to satellite data and medical images". Computer Vision and Image Understanding, vol.67, no. 3, pp. 285-295, March 1997.

[34] T. Sound, and K. Eilersten, "An Algorithm for fast adaptive binarization with application in radiotherapy imaging", Phys. Med, Biology, vol 52, pp6651-6661.

[35] Mo. Mehdizadeh, S. Dolatyarm, "Study of the Effect of Adaptive Histogram Equalization on Image Quality in Digital Preapical Image in Pre Apex Area", Reasearch Journal of Biological Sciences, vol 4, pp 922-924

[36] M.Vlachos, E. Dermatas, "Multi-scale retinal vessel segmentation using line tracking", Computerized Medical Imaging and Graphics 34 (2010) 213–227

[37] C. Sinthanayothin, J. Boyce, C.T. Williamson, Automated localisation of the optic disk, fovea, and retinal blood vessels from digital colour fundus images,Br. J. Ophthalmol. (1999) 902–910.